# Generating Optimal Plans in Highly-Dynamic Domains

**Christian Fritz** and **Sheila A. McIlraith**
Department of Computer Science,
University of Toronto,
Toronto, Ontario. Canada.
{fritz, sheila}@cs.toronto.edu

## Abstract

Generating optimal plans in highly dynamic environments is challenging. Plans are predicated on an assumed initial state, but this state can change unexpectedly during plan generation, potentially invalidating the planning effort. In this paper we make three contributions: (1) We propose a novel algorithm for generating optimal plans in settings where frequent, unexpected events interfere with planning. It is able to quickly distinguish relevant from irrelevant state changes, and to update the existing planning search tree if necessary. (2) We argue for a new criterion for evaluating plan adaptation techniques: the *relative* running time compared to the "size" of changes. This is significant since during recovery more changes may occur that need to be recovered from subsequently, and in order for this process of repeated recovery to terminate, recovery time has to *converge*. (3) We show empirically that our approach can converge and find optimal plans in environments that would ordinarily defy planning due to their high dynamics.

## 1 Introduction

A natural way for an agent to decide how to act is to exploit a policy – a function that maps each state into an action to be performed. Unfortunately, computing a policy is time intensive, so in many applications an agent plans from a known initial state instead. Unfortunately, when the application is situated within a highly dynamic environment, this initial state may rapidly change in unpredictable ways during planning, possibly invalidating the current planning effort. We argue that neither boldly ignoring such changes nor replanning from scratch is an appealing option. While the former is unlikely to produce a good plan, the latter may never be able to complete a plan when unexpected events keep interrupting. Instead we propose an integrated planning and recovery algorithm that explicitly reasons about the relevance and impact of discrepancies between assumed and observed initial state.

As a motivating example, consider a soccer playing robot in RoboCup, which, having the ball, deliberates about how to score. In RoboCup it is common to receive sensor readings 10 times per second. The game environment is very dynamic, resulting in frequent discrepancies between assumed and observed initial state. Such discrepancies may or may not affect the current planning process. But how can the robot tell? And how should the robot react when discrepancies are deemed relevant? For instance, assume that at some point during planning, the current most promising plan starts with turning slightly to face the goal and then driving there, pushing the ball. If the ball unexpectedly rolls 10 centimeters away while deliberating, the initial turn action may cause the robot to lose the ball, so this discrepancy is relevant and another plan, starting by re-approaching the ball, should be favored. But if the ball rolls closer, the original plan remains effective and the discrepancy should be ignored and planning continued.

The contributions of this paper are three-fold: (1) We propose a novel algorithm for plan generation that monitors the state of the world during planning and recovers from relevant unexpected state changes. The algorithm produces plans that are optimal with respect to the state where execution begins. It is able to distinguish between relevant and irrelevant discrepancies and updates the planning search tree to reflect the new initial state if necessary. This is generally much faster than replanning from scratch, but does not rely on knowledge about a predefined set of potential contingencies: we assume that the system can spontaneously assume any state. This is particularly interesting in continuous domains, where the number of possible discrepancies is infinite. Intuitively, the approach strikes a compromise between complete policies and simple sequential plans and uses relevance information, computed during plan generation, to increase the plan's robustness. (2) We introduce a new criterion for evaluating plan adaptation algorithms: their *relative* running time compared to the "size" of the discrepancy. We argue that this measure is of greater practical significance than either theoretical worst case considerations or the absolute recovery time. In highly dynamic domains unexpected state changes occur during planning as well as during plan adaptation. In order to obtain a plan that is known to be optimal when execution commences, the cycle of planning and recovery has to terminate by a completed recovery before the state changes any further. This is possible when the time for recovery is roughly proportional to the size of the change. Imagine planning takes 10 seconds and recovering from any state changes that occurred during that time takes 8 seconds. If we assume that in 8 seconds on average fewer changes happen than in 10, it seems reasonable to expect that we can recover from those in less than 8 seconds, say on average 6. This continues, until recovery has "caught up with reality". We informally call this behavior *convergence*. Repeated replanning from scratch does not converge, as it does not differentiate between "big" and "small" discrepancies. (3) We show empirically that our algorithm converges often even when unexpected state changes occur at relatively high frequency. Particularly "on-the-fly" recovery, i.e., recovering immediately upon discrepancy detection, has a higher chance of convergence than the alternative of completing the original planning task first and recovering only afterwards.

We explicitly assume that the number and extent of discrepancies is on average proportional to the time interval, i.e., that greater discrepancies are incurred in longer time intervals. This seems reasonable to us and holds for many interesting application domains. This, together with the observation that our algorithm can recover from a few small changes faster than from many large ones, allows our approach to converge. We demonstrate this and the resulting convergence of our approach empirically, on domain simulations which satisfy this assumption. We further understand *optimality* to be defined in terms of what is currently known, and we want to execute plans only when they are considered optimal at the moment execution begins. Our working assumption is that no model of future exogenous events exists. Hence, we treat the problem as a deterministic planning problem, whose plans and planning process we aim to make robust against potential unexpected changes.

After reviewing some preliminaries in the next section, we describe our approach in Section 3, followed by empirical results and a discussion including related work.

## 2 Background

For the exposition in this paper, we use the situation calculus with a standard notion of arithmetic, but the approach works with any action specification language for which regression can be defined, including STRIPS and ADL.

The situation calculus is a logical language for specifying and reasoning about dynamical systems [Reiter, 2001].

In the situation calculus, the *state* of the world is expressed in terms of functions and relations, called *fluents* (set $\mathcal{F}$), relativized to a *situation* $s$, e.g., $F(\vec{x}, s)$. A situation is a *history* of the primitive actions performed from a distinguished initial situation $S_0$. The function $do(a, s)$ maps an action and a situation into a new situation thus inducing a tree of situations rooted in $S_0$. For readability, action and fluent arguments are often suppressed. Also, $do(a_n, do(a_{n-1}, \ldots do(a_1, s)))$ is abbreviated to $do([a_1, \ldots, a_n], s)$ or $do(\vec{a}, s)$ and we define $do([\,], s) = s$. In this paper we distinguish between a finite set of *agent actions*, $\mathcal{A}_{\text{agent}}$, and a possibly infinite set of *exogenous actions* (or *events*), $\mathcal{A}_{\text{exog}}$, ($\mathcal{A} = \mathcal{A}_{\text{agent}} \cup \mathcal{A}_{\text{exog}}$). The agent can only perform agent actions, and exogenous events can happen at any time, including during planning.

A basic action theory in the situation calculus, $\mathcal{D}$, comprises four *domain-independent foundational axioms*, and a set of *domain-dependent axioms*. Details of the form of these axioms can be found in [Reiter, 2001]. Included in the domain-dependent axioms are the following sets:

**Initial state**: a set of first-order sentences relativized to situation $S_0$, specifying what is true in the initial state.

**Successor state axioms:** provide a parsimonious representation of frame and effect axioms under an assumption of the completeness of the axiomatization. There is one successor state axiom for each fluent, $F$, of the form $F(\vec{x}, do(a, s)) \equiv \Phi_F(\vec{x}, a, s)$, where $\Phi_F(\vec{x}, a, s)$ is a formula with free variables among $\vec{x}, a, s$. $\Phi_F(\vec{x}, a, s)$ characterizes the truth value of the fluent $F(\vec{x})$ in the situation $do(a, s)$ in terms of what is true in situation $s$. These axioms can be automatically generated from effect axioms.

**Action precondition axioms:** specify the conditions under which an action is possible. There is one axiom for each $a \in \mathcal{A}_{\text{agent}}$ of the form $Poss(a(\vec{x}), s) \equiv \Pi_a(\vec{x}, s)$ where $\Pi_a(\vec{x}, s)$ is a formula with free variables among $\vec{x}, s$. We assume exogenous events $e \in \mathcal{A}_{\text{exog}}$ are always possible.

**Regression**
The *regression* of a formula $\psi$ through an action $a$ is a formula $\psi'$ that holds prior to $a$ being performed if and only if $\psi$ holds after $a$ is performed. In the situation calculus, one step regression is defined inductively using the successor state axiom for a fluent $F(\vec{x})$ as above [Reiter, 2001]:

$$Regr[F(\vec{x}, do(a, s))] = \Phi_F(\vec{x}, a, s)$$
$$Regr[\neg \psi] = \neg Regr[\psi]$$
$$Regr[\psi_1 \wedge \psi_2] = Regr[\psi_1] \wedge Regr[\psi_2]$$
$$Regr[(\exists x)\psi] = (\exists x)Regr[\psi]$$

We use $\mathcal{R}[\psi(s), \alpha]$ to denote $Regr[\psi(do(\alpha, s))]$, and $\mathcal{R}[\psi(s), \vec{\alpha}]$ to denote the repeated regression over all actions in the sequence $\vec{\alpha}$ (in reverse order). Note that the resulting formula has a free variable $s$ of sort situation. Intuitively, it is the condition that has to hold in $s$ in order for $\psi$ to hold after executing $\vec{\alpha}$ (i.e. in $do(\vec{\alpha}, s)$). It is predominantly comprised of the fluents occurring in the conditional effects the actions in $\vec{\alpha}$. Due to the Regression Theorem [Reiter, 2001] we have that $\mathcal{D} \models \psi(do(\vec{\alpha}, s)) \equiv \mathcal{R}[\psi(s), \vec{\alpha}]$ for *all* situations $s$. Regression is therefore independent from the state where the resulting formula is evaluated.

Regression is a purely syntactic operation. Nevertheless, it is often beneficial to simplify the resulting formula for later evaluation. Regression can be defined in many action specification languages. In STRIPS, regression of a literal $l$ over an action $a$ is defined based on the add and delete lists of $a$. Regression in ADL was defined in [Pednault, 1989].

**Notation:** We use $\alpha$ to denote arbitrary but explicit actions and $S$ to denote arbitrary but explicit situations, that is $S = do(\vec{\alpha}, S_0)$ for some explicit action sequence $\vec{\alpha}$. Further $\vec{\alpha} \cdot \alpha$ denotes the result of appending action $\alpha$ to the sequence $\vec{\alpha}$.

Going back to our RoboCup example, regressing the goal "ball in goal" over the action "drive to goal", yields a condition "have ball". The further regression over a "turn" action states "distance to ball < $10 cm$" as a condition for the success of the considered plan, if, e.g., the robot's 10cm long grippers enable turning with the ball.

## 3 Planning with Unexpected Events

In this paper we consider a planner based on $A^*$ forward search that uses positive action costs as a metric, but the conceptual approach is amenable to a variety of other forward-search based planning techniques and paradigms.

Intuitively, our approach annotates the search tree with all relevant information for determining the optimal plan. By regressing the goal, preconditions, and metric function over all considered action sequences, this information is expressed in terms of the current state. When unexpected events change the current state of the world, this allows us to reason symbolically about their relevance and their potential impact on the current search tree and choice of plan—much faster than replanning from scratch.

For instance, our soccer robot from above knows from regressing the goal that the plan ["turn", "drive to goal"] will succeed whenever "distance to ball < $10cm$" holds. Hence it can determine the relevance of the aforementioned ball displacements, and also that, for instance, unexpected actions of its teammates can be ignored for now. A complication of this arises from our interest in optimal, rather than just valid plans, however. We will need to also consider alternative action sequences, and also handle impacts on the regressed metric function.

At the highest level, the approach we present here consists of two components: A regression-based $A^*$ planner, and a recovery procedure. These components, described below, can be used in at least two possible ways:

**At-the-end:** The planner generates an optimal plan for the assumed initial state. If no changes to the initial state occur, the resulting plan is optimal and execution can commence. Otherwise, the recovery procedure updates the final search tree and open list (describing the search frontier) as necessary given any observed changes to the initial state. If the order of the open list does not change during the recovery, and hence still has the previously found plan as its first element, the plan is known to remain optimal and can be executed. Otherwise, the planner resumes plan generation given the updated structures.

**On-the-fly:** In the absence of any changes to the assumed initial state, the planner is proceeding in its search for an optimal plan. Whenever a change to the initial state is observed, plan generation is interrupted and the recovery procedure updates the current search tree and open list to reflect the changes. Plan generation then continues.

In both cases, this alternation continues until a plan generation cycle terminates without further interruptions, in which case the resulting plan is known to be optimal with respect to the currently assumed initial state. Note that the following descriptions are simplified for readability. In practice, there are various ways for increasing the efficiency of the implementation.

### 3.1 Regression-based $A^*$ planning

In this section we present an $A^*$ planner that returns not only a plan, but also the remaining open list upon termination of search, as well as a search tree annotated with any relevant regressed formulae.

To provide a formal characterization, we assume that the planning domain is encoded in a basic action theory $\mathcal{D}$. Given an initial situation $S$, a goal formula $Goal(s)$, a formula $Cost(a, c, s)$ defining costs $c$ of action $a$, and a consistent heuristic specified as a formula $Heu(h, s)$, $A^*$ search finds a sequence of actions $\vec{\alpha}$ such that the situation $do(\vec{\alpha}, S)$ satisfies the goal while minimizing the accumulated costs. Starting with an open list, *Open*, containing only one element representing the empty action sequence, the search proceeds by repeatedly removing and expanding the first element from the list until that element satisfies the goal, always maintaining the open list's order according to the value $v$ defined by:

$$\begin{aligned}
Value\bigl(v, do([\alpha_1, \ldots, \alpha_N], s)\bigr) \stackrel{\text{def}}{=} \\
(\exists h, c_1, \ldots, c_N). v = h + c_1 + \cdots + c_N \\
\land\, Heu\bigl(h, do([\alpha_1, \ldots, \alpha_N], s)\bigr) \land Cost(\alpha_1, c_1, s) \\
\land \cdots \land Cost\bigl(\alpha_N, c_N, do([\alpha_1, \ldots, \alpha_{N-1}], s)\bigr)
\end{aligned}$$

We assume that the goal can only be achieved by a particular agent action *finish*. Any planning problem can be transformed to conform to this by defining the preconditions of *finish* according to the original goal.

Our regression-based version of $A^*$ is shown in Figure 1. It interacts with the basic action theory $\mathcal{D}$ to reason about the truth-values of formulae. We say $\psi$ holds, to mean that it is entailed by $\mathcal{D}$. The algorithm is initially invoked as

```
regrA*(D, S, Goal, Cost, Heu, Open, T) :
1  if Open = [ ] then return ([ ], T)
2  else [(g, h, α⃗) | Open'] = Open        // slice first element
3  if Goal(do(α⃗, S)) holds then return (Open, T)
4  else foreach α' ∈ A_agent do
5      α⃗' ← α⃗ · α'                       // append action to sequence
6      T(α⃗').P(s) ← R[Poss(α', s), α⃗]
7      if T(α⃗').P(S) holds then
8          T(α⃗').p ← true                 // action currently possible
9          T(α⃗').C(c, s) ← R[Cost(α', c, s), α⃗]
10         T(α⃗').H(h, s) ← R[Heu(h, s), α⃗']
11         T(α⃗').c ← c' with c' s.t. T(α⃗').C(c', S) holds
12         T(α⃗').h ← h' with h' s.t. T(α⃗').H(h', S) holds
13         insert (g + c', h', α⃗') into Open'
14     else T(α⃗').p ← false              // action currently impossible
15 return regrA*(D, S, Goal, Cost, Heu, Open', T)
```

Figure 1: Pseudo-code for regression-based $A^*$ planning.

$regrA^*(\mathcal{D}, S, Goal, Cost, Heu, [(0, \infty, [])], nil)$. The last argument denotes a data structure representing the annotated search tree and is initially empty. The elements of the open list are tuples $(g, h, \vec{\alpha})$, where $\vec{\alpha} = [\alpha_1, \ldots, \alpha_n]$ is an action sequence, $g$ are the costs accumulated when executing this sequence in $S$, and $h$ is the value s.t. $Heu(h, do(\vec{\alpha}, S))$ holds. When an element is expanded, it is removed from the open list and the following is performed for each agent action $\alpha'$: First, the preconditions of $\alpha'$ are regressed over $\vec{\alpha}$ (Line 6). If the resulting formula, stored in $T(\vec{\alpha}).P(s)$, holds in $S$ according to $\mathcal{D}$ (Line 7), the cost formula for $\alpha'$ is regressed over $\vec{\alpha}$, the heuristic is regressed over $\vec{\alpha} \cdot \alpha'$, and the resulting formulae are evaluated in $S$ yielding values $c'$ and $h'$ (Lines 9–12). Intuitively, the regression of these formulae over $\vec{\alpha}$ describes, in terms of the current situation, the values they will take after performing $\vec{\alpha}$. Finally, a new tuple is inserted into the open list (Line 13) according to $g + c' + h'$ to maintain the open list's order according to $Value(v, s)$.

$A^*$ keeps expanding the first element of the open list until this element satisfies the goal, in which case the respective action sequence describes an optimal plan. This is because a consistent heuristic never over-estimates the actual remaining costs from any given state to the goal. Due to the Regression Theorem [Reiter, 2001], this known fact about $A^*$ also holds for our regression-based version. Similarly the completeness of $A^*$ is preserved.

In service of our recovery algorithm described below, we explicitly keep the search tree $T$ and annotate its nodes with the regressed formulae for preconditions $(T(\vec{\alpha}).P(s))$, costs $(T(\vec{\alpha}).C(c, s))$, and heuristic value $(T(\vec{\alpha}).H(h, s))$ and their values according to the (current) initial situation $S$ $(T(\vec{\alpha}).p, T(\vec{\alpha}).c,$ and $T(\vec{\alpha}).h)$. Roughly, when this state changes due to an unexpected event $e$, we reevaluate $T(\vec{\alpha}).P(s), T(\vec{\alpha}).C(c, s),$ and $T(\vec{\alpha}).H(h, s)$ in $s = do(e, S)$, and update their values and the open list accordingly. However, not all of these reevaluations are actually necessary and this is the *key insight* providing the speed-up of our

```
recover(D, S_1, S_2, Cost, Heu, Open, T, Index) :
1  F_Δ ← {F ∈ keys(Index) | F(S_1) ≢ F(S_2) holds }
2  Δ ← ⋃_{F ∈ F_Δ} Index(F)              // affected formulae
3  foreach (α⃗, 'p') ∈ Δ do               // update preconditions
4      if T(α⃗).p = true and ¬T(α⃗).P(S_2) holds then
5          T(α⃗).p ← false                // action now impossible
6          foreach (g, h, α⃗') ∈ Open do
7              if α⃗ is prefix of α⃗' then
8                  remove (g, h, α⃗') from Open
9      elseif T(α⃗).p = false and T(α⃗).P(S_2) holds then
10         T(α⃗).p ← true                 // action now possible
11         α⃗' · α_last = α⃗              // get last action in sequence
12         T(α⃗).C(c, s) ← R[Cost(α_last, c), α⃗']
13         T(α⃗).H(h, s) ← R[Heu(h), α⃗']
14         T(α⃗).c ← c' with c' s.t. T(α⃗).C(c', S_2) holds
15         T(α⃗).h ← h' with h' s.t. T(α⃗).H(h', S_2) holds
16         g' ← getGval(T, α⃗)
17         insert (g', h', α⃗) into Open and update Index
18 foreach (α⃗, 'c') ∈ Δ do                // update accumulated costs
19     get c' s.t. T(α⃗).C(c', S_2) holds
20     offset ← c' − T(α⃗).c
21     foreach (g, h, α⃗') ∈ Open do
22         if α⃗ is prefix of α⃗' then g ← g + offset
23     T(α⃗).c ← c'
24 foreach (α⃗, 'h') ∈ Δ do                // update heuristic values
25     if (∃g, h).(g, h, α⃗) ∈ Open then
26         h ← h' with h' s.t. T(α⃗).H(h', S_2) holds
27         T(α⃗).h ← h
28 return (sort(Open), T)
```

Figure 2: Pseudo-code of our recovery algorithm.

algorithm: since all formulae are regressed and hence expressed in terms of the current state, we can determine which ones are actually affected by the event, by simply considering the fluents the formulae mention. For this purpose, we maintain an index *Index* whose keys are ground fluents (e.g., *distanceTo(ball)*) and whose values are lists of pointers to all stored formulae that mention it.

### 3.2 Recovering from Unexpected Events

While generating a plan for an assumed initial situation $S$, an unexpected event $e$, say "*distanceTo(ball)* ← 20", may occur, changing the state of the world and putting us into situation $do(e, S)$. When this happens, the aforementioned index lets us pinpoint all formulae affected by this change (e.g., $T([turn, driveTo(goal), finish]).P(s))$. After reevaluating these formulae in $do(e, S)$ and updating their values, the search tree will be up-to-date in the sense that all contained values are with respect to $do(e, S)$ rather than the originally assumed situation $S$. After propagating these changes to the open list, search can continue. We show that the resulting plan is optimal for the new situation. *Note that the regressed formulae never change.* Since often only very few fluents are affected by unexpected events, this relevance-based approach allows for very efficient recovery.

The recovery algorithm is specified in Figure 2. $T$ denotes the annotated search tree, *Open* is the open list, and *Index*

the index. The latter contains entries of the form $(\vec{\alpha}, type)$, where $\vec{\alpha}$ is a sequence of actions and $type$ is either of 'p', 'c', or 'h'. The algorithm modifies the values of the tree and the open list (ll. 22 and 26) to reflect their value with respect to a new situation $S_2$ (e.g., $do(e, S_1)$) rather than an originally assumed initial situation $S_1$. If the event changes the truth value of action preconditions, the content of the open list is modified accordingly (ll. 8, 17). When a previously impossible action has now become possible (Line 9) the annotation for this node is created and a new entry added to the open list (ll. 11-17). The function $getGval(T, \vec{\alpha})$ computes the sum of all costs $(T(\cdot).c)$ annotated in $T$ along the branch from the root to node $\vec{\alpha}$.

As mentioned before, the algorithm can be used in one of at least two ways: on-the-fly, dealing with unexpected state changes immediately, or at-the-end, dealing at once with all events that occurred during planning. The former has the advantage that the planning effort is focused more tightly on what is actually relevant given everything that has happened so far. This approach can be implemented by inserting code right before Line 15 of *regrA\** that checks for events and invokes *recover* if necessary, changing $S, Open'$, and $T$ accordingly. The appeal of the latter stems from the fact that recovering from a bulk of events simultaneously can be more efficient than recovering from each individually. It may, however, be necessary to resume *regrA\** search afterwards, if, for instance, the current plan is no longer valid in the new initial state or a new opportunity exists, which may lead to a better plan. With both approaches, additional events may happen during recovery, making additional subsequent recoveries necessary.

The following theorem states the correctness of *recover* in terms of the at-the-end approach: calling *recover* and continuing *regrA\** with the new open list, produces an optimal plan and in particular the same as replanning from scratch in $S_2$. Recall that the head of the open list contains the optimal plan. For on-the-fly, correctness can be shown analogously (cf. Lemma 1 in [Fritz, 2009, p.191]).

**Theorem 1** (Correctness). Let $\mathcal{D}$ be a basic action theory, *Goal* a goal formula, $Cost(a, c)$ a cost formula, and $Heu(h)$ a consistent heuristic. Then, for any two situations $S_1, S_2$ in $\mathcal{D}$ we have that after the sequence of invocations:

1. $(O_1, T_1) \leftarrow regrA^*(\mathcal{D}, S_1, Goal, Cost, Heu, [(0, \infty, [\,])], nil)$,
2. create *Index* from $T_1$,
3. $(O_2, T_2) \leftarrow recover(\mathcal{D}, S_1, S_2, O_1, T_1, Index)$,
4. $(O_3, T_3) \leftarrow regrA^*(\mathcal{D}, S_2, Goal, Cost, Heu, O_2, T_2)$,

the first element of $O_3$ will be the same as in $O'$ of
$(O', T') \leftarrow regrA^*(\mathcal{D}, S_2, Goal, Cost, Heu, [(0, \infty, [\,])], nil)$,
or both $O_3$ and $O'$ are empty. *Proof:* [Fritz, 2009, p.190 ff.].

As a special case, this works for $S_2 = do(\vec{e}, S_1)$, for any situation $S_1$ and sequence of events $\vec{e}$. Note that such events can produce arbitrary changes to the state of the world. The algorithm does not make any assumptions about possible

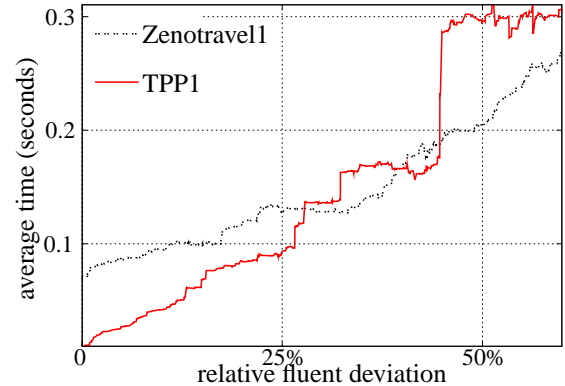

Figure 3: Recovery time relative to amount of change.

events. Any fluent may assume any value at any time.

In complex domains, many state changes are completely irrelevant to the current planning problem, overall or at the current stage of planning, and others only affect a small subset of elements in the search tree. During recovery, we exploit this structure to gain significant speed-ups compared to replanning from scratch. More importantly it allows us to recover from small perturbations faster than from large ones, where "large" may refer to the number of fluents that changed or the amount by which continuous fluents changed (cf. Section 4). This way, recovery can *converge*, i.e., "catch up with reality", as we defined informally in the introduction. We verified this empirically.

## 4 Empirical Results

We present empirical results obtained using a current implementation of our algorithm to generate optimal plans for differently sized problems of the metric TPP and Zenotravel domains of the International Planning Competition. We begin by showing that the time required for recovering from unexpected state changes is roughly and on average proportional to the extent of the change. We then show that our approach is able to find optimal plans even when the initial state changes frequently. We compare the two mentioned recovery strategies on-the-fly and at-the-end, showing that the former clearly outperforms the latter in terms of likelihood of convergence. Finally, and not surprisingly, we show that our approach generally outperforms replanning from scratch. All experiments were run on an Intel Xeon 2.66 GHz with 2GB RAM.

Figure 3 plots the average time the combination of *recover* + continued *regrA\** search took to find a new optimal plan, after the value of a randomly selected continuous fluent was randomly modified after generating an optimal plan. A deviation $x\%$ means that the fluent was multiplied by $1 \pm \frac{x}{100}$. Note that we used continuous fluents in our experiments only because they lend themselves better to a quantitative evaluation—our approach is equally applicable to discrep-ancies on discrete valued, including Boolean, fluents. As one can see, the time to recover from a drastic change takes on average longer than for minor deviations. While this doesn't seem surprising, it is a necessary condition for the convergence of the recovery, which we study next.

We assume that in shorter periods of time fewer things change or changes are less drastic than over longer periods. As we will see, recovery generally takes less time than the original planning did. Hence, we expect fewer or less drastic changes during recovery than during planning. A second recovery—from the events that occurred during the first recovery—is thus predicted to take less time than the first. This process often continues until convergence. We studied the conditions under which our algorithm converges by simulating domains with frequent changes to the initial state. At high frequencies during planning and subsequent recoveries, we randomly perturbed some fluent by an amount of up to a certain maximum between 5-80%. We considered the two approaches described earlier: completing the original planning task and recovering only afterwards followed by further $regrA^*$ search if needed (at-the-end), or reacting to state changes immediately pausing further $regrA^*$ expansion until *recover* has brought the current search tree up-to-date (on-the-fly). In both cases, several episodes of recovery and additional $regrA^*$ search were generally required before finding an optimal and up-to-date plan. Their number varied strongly, as a result of some discrepancies having larger impact than others. Table 1 shows, for different frequencies and amounts of deviation, the percentages of simulations in which an optimal plan was found, i.e., the algorithm converged within the time limit. As time limit we used 30 times the time required for solving the respective original problems, without perturbations, using a conventional $A^*$ search planner. These were 0.52s for TPP1, 2.17s for TPP2, 3.03s for TPP3, and 0.34s, 0.82s, and 1.58s for Zenotravel 1, 2, and 3 respectively. The frequencies shown in the table are relative to these as well. For instance, the value 100 for Zenotravel1 on-the-fly, 5Hz, 40% states that even when every $0.34s/5 = 68$ ms the value of a random fluent changed by up to 40% in the considered Zenotravel problem, the on-the-fly approach still converged 100% of the time. We believe this simulates a quite erratic environment, possibly harsher than many realistic application domains.

The on-the-fly recovery strategy clearly outperforms at-the-end recovery. This makes intuitive sense, as no time is wasted continuing planning for an already flawed instance. This also motivates an integrated approach, showing its benefit over the use of plan adaptation approaches which are only applicable once a first plan has been produced. The table also shows that convergence was much better on TPP than on Zenotravel. Interestingly, this was predictable given Figure 3: since the curve for Zenotravel1 intersects the y-axis at around 0.07 seconds, it seems un-reasonable to expect convergence on this problem when the initial state changes at intervals shorter than that. This explains the low probability of convergence when events occur at 10Hz times planning time, i.e., every 0.034s. In comparison, recovering via replanning from scratch takes the same amount of time, no matter how small the discrepancy is, unless the problem gets significantly easier due to the change. Hence, it has no chance of ever catching up with reality when events happen at time intervals shorter than the time required for plan generation.

Not surprisingly, our approach generally outperforms replanning from scratch. To demonstrate this, we compared the times required by both approaches for recovering from a single change of the initial state. The setup was as follows: We solved a planning problem, perturbed the state of the world by randomly changing some fluent's value, and then ran both (a) *recover* followed by further $regrA^*$ search based on the modified open list if necessary, and (b) replanning from scratch using a conventional $A^*$ search implementation using the same heuristic. The fluent perturbations were done on continuous fluents only, and the amount of change was up to 50%.

Figure 4 shows the time both approaches require to recover from single events on our TPP1 problem. Recall that with both approaches the resulting plan is provably optimal. We separately show the times for cases where (a) additional $regrA^*$ was necessary, and (b) where it was not. The latter is the case when following the *recover*, the first element of the open list satisfied the goal. The average speed-up over replanning from scratch was 10.56 in the former case, and 33.64 in the latter. In all test cases, the simulated discrepancy was relevant in the sense that at least one formula appearing in the annotation was affected. Hence, calling *recover* was necessary in all cases—the set $\Delta$ of Figure 2 was never empty.

We performed the same experiment on the Zenotravel1 problem. It is a reasonable question to ask whether the relative speed-up of our approach is just due to the use of a comparatively slow replanner. Therefore we tested using two different, hand-coded heuristics, where the first is more informed (i.e., better) than the second. Using the first, which we also used in the earlier described experiments, the average recovery time was 0.14s, and the average replanning time was 0.51s, whereas with the second heuristic recovery time averaged to 0.35s and replanning to 1.07s. This shows that even when the planner, and thus replanner, is improved by the use of a better heuristic, our approach is still generally superior to replanning from scratch. This is because it equally benefits from a smaller search tree, resulting from the use of a better heuristic.

Unfortunately, there are no other replanning systems we can compare our algorithm to, since no other system aims to produce optimal plans. Further, even for recent validity-only replanners, we were unable to obtain implementations from the authors. In addition, such replanners could only be used for at-the-end recovery, which seems inferior to on-the-fly recovery.

| Frequency: | 3Hz · planning time | | | | | 5Hz · planning time | | | | | 10Hz · planning time | | | | |
|---|---|---|---|---|---|---|---|---|---|---|---|---|---|---|---|
| Deviation: | 5% | 10% | 20% | 40% | 80% | 5% | 10% | 20% | 40% | 80% | 5% | 10% | 20% | 40% | 80% |
| tpp1 at-the-end | 100 | 100 | 100 | 83 | 60 | 100 | 100 | 83 | 63 | 43 | 100 | 100 | 76 | 43 | 20 |
| tpp1 on-the-fly | 100 | 100 | 100 | 86 | 83 | 100 | 100 | 96 | 80 | 83 | 100 | 100 | 93 | 80 | 70 |
| tpp2 at-the-end | 96 | 86 | 60 | 63 | 43 | 100 | 80 | 51 | 44 | 34 | 89 | 48 | 34 | 24 | 10 |
| tpp2 on-the-fly | 100 | 93 | 86 | 86 | 83 | 96 | 86 | 75 | 86 | 82 | 96 | 86 | 86 | 79 | 82 |
| tpp3 at-the-end | 100 | 73 | 50 | 66 | 41 | 94 | 72 | 55 | 42 | 52 | 76 | 31 | 42 | 32 | 20 |
| tpp3 on-the-fly | 100 | 96 | 87 | 92 | 72 | 94 | 84 | 86 | 87 | 89 | 89 | 81 | 81 | 89 | 86 |
| zenotravel1 at-the-end | 100 | 96 | 100 | 100 | 76 | 66 | 76 | 63 | 43 | 56 | 3 | 6 | 0 | 6 | 16 |
| zenotravel1 on-the-fly | 100 | 100 | 100 | 100 | 100 | 96 | 96 | 100 | 100 | 86 | 66 | 73 | 70 | 93 | 93 |
| zenotravel2 at-the-end | 66 | 43 | 30 | 26 | 3 | 30 | 16 | 6 | 6 | 6 | 10 | 0 | 0 | 0 | 0 |
| zenotravel2 on-the-fly | 86 | 70 | 53 | 53 | 40 | 36 | 16 | 30 | 26 | 23 | 13 | 6 | 0 | 6 | 20 |
| zenotravel3 at-the-end | 100 | 80 | 56 | 28 | 8 | 97 | 72 | 12 | 7 | 7 | 33 | 10 | 0 | 0 | 2 |
| zenotravel3 on-the-fly | 100 | 92 | 60 | 56 | 66 | 90 | 75 | 60 | 30 | 43 | 43 | 28 | 33 | 25 | 21 |

Table 1: Percentage of test cases where our approach converged within the time limit, by event frequencies and deviation amounts.

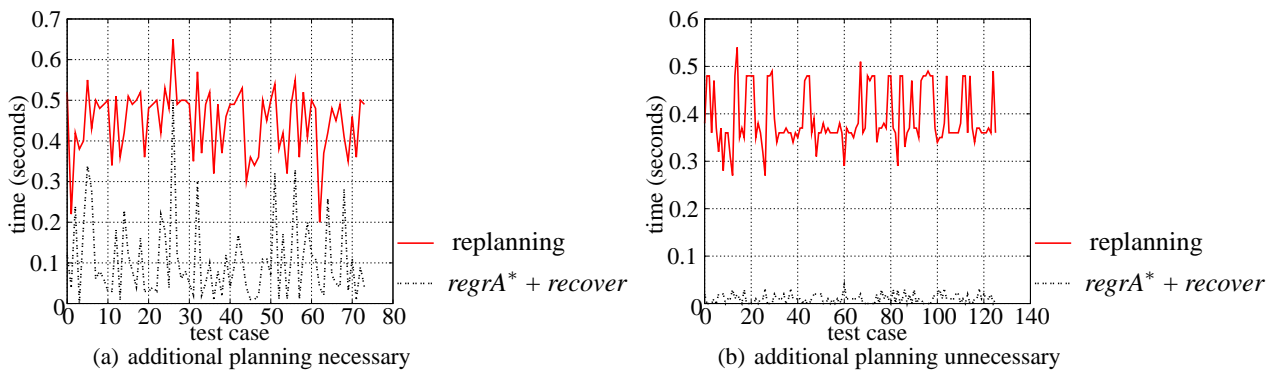

(a) additional planning necessary  (b) additional planning unnecessary

Figure 4: Run-time comparison of our approach vs. replanning from scratch on the TPP domain.

## 5 Discussion

We made three contributions: (1) We presented a novel integrated planning and recovery algorithm for generating optimal plans in environments where the state of the world frequently changes unexpectedly during planning. At its core, the algorithm reasons about the relevance and impact of discrepancies, allowing the algorithm to recover from changes more efficiently than replanning from scratch. (2) We introduced a new criterion for evaluating plan adaptation approaches, called convergence, and argued for its significance. (3) We provided empirical evidence for the convergence of our approach under high frequencies of unexpected state changes. Our experiments also show that an interleaved planning+recovery approach which recovers from discrepancies on-the-fly is superior to an approach that only recovers once planning has completed.

In the future, we intend to apply this work to a highly dynamic real-world domain such as the mentioned RoboCup or Unmanned Aerial Vehicles. To do so, an optimized version of our implementation is required. While the approach is able to handle changes in the state, which can also be used to model changes in executability and cost of actions, we would like to study changing goals as well. We also think that the ideas behind the presented approach may be beneficially applied to planning under initial state uncertainty, in particular when such uncertainty ranges over continuous domains.

The presented work is part of a larger body of research regarding the generation and execution of optimal plans in highly dynamic domains (cf. [Fritz, 2009]), including an approach for monitoring plan optimality during execution. This is needed since during execution the optimality of the executing plan may also be jeopardized by exogenous events.

The presented approach is one of the first to monitor and react to unexpected state changes during planning. The approach taken by Veloso et al. [1998] exploits the "rationale", the reasons for choices made during planning, to deal with discrepancies that occur during planning. They acknowledge the possibility that previously sub-optimal alternatives may become better than the current plan candidate as the world evolves during planning, but the treatment of optimality is informal and limited. No guarantees are made regarding the optimality of the resulting plan. Also, by us-ing best-first search, our approach is compatible with many state-of-the-art planners, while the approach of Veloso *et al.* is particular to the PRODIGY planner.

Several approaches exist for adapting a plan in response to unexpected events that occur during execution, rather than during plan generation, e.g., [Koenig *et al.*, 2002; Hanks and Weld, 1995; Gerevini and Serina, 2000]. Arguably we could use these approaches for our purpose of recovering from discrepancies during planning by first ignoring the changes and then recovering once a plan is generated. We think this is inferior to our approach for the following reasons: (1) except for the first, the listed approaches do not guarantee optimality, (2) we have shown that an integrated approach which recovers from state changes on-the-fly has convergence advantages, and (3) it is not clear whether such a use of these replanners would at all lead to convergence.

The SHERPA system presented by Koenig *et al.* [2002] monitors the continued optimality of a plan only in a limited form. SHERPA lifts the Life-Long Planning A$^*$ (LPA$^*$) search algorithm to symbolic propositional planning. LPA$^*$ was developed for the purpose of replanning in problems like robot navigation (i.e., path replanning) with simple, unconditional actions, and only applies to replanning problems where the costs of actions have changed but the current state remains the same. Similar to our approach, SHERPA retains the search tree to determine how changes may affect the current plan. Our approach subsumes this approach and further allows for the general case where the initial (current) state may change arbitrarily and the dynamics of the domain may involve complex conditional effects. SHERPA's limitations equally apply to more recent work by Sun and Koenig [2007]. The presented Fringe-Saving A$^*$ (FSA$^*$) search algorithm, which sometimes performs better than LPA$^*$, is further limited to grid world applications and the use of the Manhattan distance heuristic. This algorithm retains the open list of previous searches as well.

The idea of deriving and utilizing knowledge about relevant conditions of the current state for monitoring and possibly repairing a plan, has been used before, e.g., Kambhampati [1990], and reaches back to the early work on Shakey the Robot by Fikes *et al.* [1972]. Fikes *et al.* used triangle tables to annotate the plan with the regressed goal, in order to determine whether replanning was necessary when the state of the world changed unexpectedly.

Nebel and Koehler [1995] show that plan reuse, and hence plan repair, has the same worst case complexity as planning from scratch. This result is interesting in theory, but not so relevant in the practical case of optimal plan generation in the face of frequent unexpected events. In this case, we have shown that if we want to have a plan that we know to be optimal at the start of execution, then the recovery time relative to the impact of an event is more important.

An important merit of our approach is its ability to handle any possible state change, rather than being limited to a predefined set of contingencies, but without paying the price of computing a complete policy. It hence explores a pragmatic middle-ground between the efficiency of deterministic planning, and the robustness of solving MDPs completely. The lack of the explicit consideration of possible contingencies during planning, distinguishes this work from related approaches for solving relational and first-order MDPs (e.g., [Boutilier *et al.*, 2001; Hölldobler *et al.*, 2006]).

**Acknowledgements:**
We gratefully acknowledge funding from the Natural Sciences and Engineering Research Council of Canada.